%% file: main.tex
\documentclass[11pt,a4paper]{article}
\usepackage[hyperref]{emnlp2020}
\usepackage{times}
\usepackage{latexsym}
\usepackage{graphicx}
\usepackage{subcaption}
\usepackage[hide]{notes-alt}

\aclfinalcopy

\newcommand\bert{\textsc{BERT}}
\newcommand\mlmsquad{\textsc{Mlm-SQuAD}}
\newcommand\mlmmarco{\textsc{Mlm-MSMarco}}
\newcommand\qasquad{\textsc{Qa-SQuAD-1}}
\newcommand\qasquadbig{\textsc{Qa-SQuAD-2}}

\newcommand\rankmarco{\textsc{Rank-MSMarco}}
\newcommand\ner{\textsc{Ner-CoNLL}}
\newcommand\squadprobe{\texttt{Squad}}
\newcommand\trexprobe{\texttt{T-REx}}
\newcommand\concpetprobe{\texttt{ConceptNet}}
\newcommand\greprobe{\texttt{Google-RE}}
\newcommand{\mpara}[1]{\medskip\noindent{\bf #1}}

\title{BERTnesia: Investigating the capture and forgetting of knowledge in BERT}

\author{Jonas Wallat \\
 L3S Research Center \\
 Hannover, Germany \\
 \texttt{wallat@l3s.de} \\\And
 Jaspreet Singh \\
 L3S Research Center \\
 Hannover, Germany \\
 \texttt{singh@l3s.de} \\
 \And
 Avishek Anand \\
 L3S Research Center \\
 Hannover, Germany \\
 \texttt{anand@l3s.de} \\}

\date{}

\begin{document}
\maketitle
\begin{abstract}
Probing complex language models has recently revealed several insights into linguistic and semantic patterns found in the learned representations. In this paper, we probe BERT specifically to understand and measure the relational knowledge it captures. We utilize knowledge base completion tasks to probe every layer of pre-trained as well as fine-tuned BERT (ranking, question answering, NER). 
Our findings show that knowledge is not just contained in BERT’s final layers. Intermediate layers contribute a significant amount (17-60\%) to the total knowledge found. Probing intermediate layers also reveals how different types of knowledge emerge at varying rates. When BERT is fine-tuned, relational knowledge is forgotten but the extent of forgetting is impacted by the fine-tuning objective but not the size of the dataset. We found that ranking models forget the least and retain more knowledge in their final layer. We release our code on github\footnote{https://github.com/jwallat/knowledge-probing} to repeat the experiments.

\end{abstract}

\input{intro-avi}

\input{rel-work}

\input{setup}

\input{evolution-new}

\input{fine-tuned}

\input{conclusion}

\newpage

\bibliography{anthology,new_bib}
\bibliographystyle{acl_natbib}

\newpage

\input{appendix}



\end{document}

%% file: intro-avi.tex
\section{Introduction}

Large pre-trained language models like \bert{}~\cite{devlin2018bert} have heralded an \textit{Imagenet} moment for NLP\footnote{https://thegradient.pub/nlp-imagenet/} with not only significant improvements made to traditional tasks such as question answering and machine translation but also in the new areas such as knowledge base completion. 
\bert{} has over 100 million parameters and essentially trades off transparency and interpretability for performance. Loosely speaking, probing is a commonly used technique to better understand the inner workings of \bert{} and other complex language models~\cite{dasgupta2018evaluating,ettinger-etal-2018-assessing}. 
Probing, in general, is a procedure by which one tests for a specific pattern -- like local syntax, long-range semantics or even compositional reasoning -- by constructing inputs whose expected output would not be possible to predict without the ability to detect that pattern. 
While a large body of work exists on probing \bert{} for linguistic patterns and semantics, there is limited work on probing these models for the factual and relational knowledge they store.

Recently, \citet{petroni2019language} probed \bert{} and other language models for relational knowledge (e.g., \textit{Trump} \textsf{is the president of} the \textit{USA}) in order to determine the potential of using language models as automatic knowledge bases. 
Their approach converted queries in the knowledge base (KB) completion task of predicting arguments or relations from a KB triple into a natural language cloze task, e.g., \texttt{[MASK]} \textsf{is the president of} the \textit{USA}.
This is done to make the query compatible with the pre-training masked language modeling (MLM) objective. They consequently showed that a reasonable amount of knowledge is captured in \bert{} by considering multiple relation probes. 
However, there are some natural questions that arise from these promising investigations:
\textit{Is there more knowledge in \bert{} than what is reported? What happens to relational knowledge when \bert{} is fine-tuned for other tasks? Is knowledge gained and lost through the layers?}

\mpara{Our Contribution.} In this paper, we study the emergence of knowledge through the layers in \bert{} by devising a procedure to estimate knowledge contained in every layer and not just the last (as done by~\citet{petroni2019language}). 
While this type of layer-by-layer probing has been conducted for syntactic, grammatical, and semantic patterns; knowledge probing has only been conducted on final layer representations. 
Observing only the final layer (as we will show in our experiments)  (i)  underestimates the amount of knowledge and (ii) does not reveal how knowledge emerges. 
Furthermore, we explore how knowledge is impacted when fine-tuning on knowledge-intensive tasks such as question answering and ranking. 
We list the key research questions we investigated and key findings corresponding to them:

\mpara{RQ I: Do intermediary layers capture knowledge not present in the last layer?} (Section~\ref{sec:intermediate})
  \paragraph{}   
     We find that a substantial amount of knowledge is stored in the intermediate layers ($\approx24\%$ on average)


\mpara{RQ II: Does all knowledge emerge at the same rate? Do certain types of relational knowledge emerge more rapidly?} (Section~\ref{sec:evolution})
\paragraph{}
        We find that not all relational knowledge is captured gradually through the layers with 15\% of relationship types essentially doubling in the last layer and 7\% of relationship types being maximally captured in an intermediate layer. 


\mpara{RQ III: What is the impact of  fine-tuning data on knowledge capture?} (Section~\ref{sec:data_size})
\paragraph{}
        We find that the dataset size does not play a major role when the training objective is fixed as MLM. Fine-tuning on a larger dataset does not lead to less forgetting.


\mpara{RQ IV: What is the impact of the fine tuning objective on knowledge capture ? } (Section~\ref{sec:finetuning})
\paragraph{}
        Fine tuning always causes forgetting. When the size of the dataset is fixed and training objective varies, the ranking model (\rankmarco{} in our experiments) forgets less than the QA model.  


\todo{maybe write a line or two about the implications....use of \bert{} for KB completion, use of early exit models, etc etc.}

%% file: rel-work.tex
\section{Related Work}

In this section, we survey previous work on probing language models (LMs) with a particular focus on contextual embeddings learned by \bert{}. Probes have been designed for both static and contextualized word representations. Static embeddings refer to non-contextual embeddings such as GloVe~\cite{pennington2014glove}. 
For the static case, the reader can refer to this survey by ~\citet{Belinkov_2019}. 
Now we detail probing tasks for contextualized embeddings from language models.



\subsection{Probing for syntax, semantics, and grammar}


Initial work on probing dealt with linguistic pattern detection. 
\citet{Peters2018DissectingCW} investigated the ability of various neural network architectures that learn contextualized word representations to capture local syntax and long-range semantics like co-reference resolution while
\citet{dasgupta2018evaluating,ettinger-etal-2018-assessing} probed language models for compositional reasoning. 


\citet{McCoy_2019, goldberg2019assessing} found that \bert{} is able to effectively learn syntactic heuristics with natural language inference specific probes.  \citet{tenney2019bert, liu2019linguistic, jawahar-etal-2019-bert} investigated \bert{} layer-by-layer for various syntactic and semantic patterns like part-of-speech, named entity recognition, co-reference resolution, entity type prediction, semantic role labeling, etc.
They all found that basic linguistic patterns like part of speech emerge at the lower layers.
However, there is no consensus with regards to semantics with somewhat conflicting findings (equally spread vs final layer~\cite{jawahar-etal-2019-bert}). 
~\citet{kovaleva-etal-2019-revealing} found that the last layers of fine-tuned \bert{} contain the most amount of task-specific knowledge. \citet{vanaken_2019} showed the same result for fined tuned QA \bert{} with specially designed probes. They found that the lower and intermediary layers were better suited to linguistic subtasks associated with QA.
For a comprehensive survey we point the reader to~\cite{rogers2020primer}



Our work is similar to these studies in terms of setup. 
In particular, our probes function on the sentence level and are applied to each layer of a pre-trained \bert{} model as well as \bert{} fine-tuned on several tasks. 
However, we do not focus on detecting linguistic patterns and focus on relational and factual knowledge.



\subsection{Probing for knowledge}
\label{sec:know_probe}

In parallel, there have been investigations into probing for factual and world knowledge. 
Most recently, \citet{petroni2019language} found that LMs like \bert{} can be directly used for the task of knowledge base completion since they are able to memorize more facts than some automatic knowledge bases. 
They created cloze statement tasks for factual and commonsense knowledge and measured cloze-task performance as a proxy for the knowledge contained. 
However, using the same probing framework, \citet{kassner2020negated} showed that this factoid knowledge is influenced by surface-level stereotypes of words. For example, \bert{} often predicts a typically German name as a German citizen. Tangentially, \citet{forbes2019neural} investigated \bert{}'s awareness of the world. They devised object property and action probes to estimate \bert{}'s ability to reason about the physical world. 
They found that \bert{} is relatively incapable of such reasoning but is able to memorize some properties of real-world objects. This investigation tested common sense spatial reasoning rather than pure factoid knowledge.

Rather than focusing on newer knowledge types, we focus on the true coverage of already known relations and facts in~\bert{}. In terms of experiments, we do not focus on knowledge containment in different language models, rather focus on investigating how knowledge emerges specifically in \bert{}. Here, we are more interested in relative differences. To this end, we devise a procedure to adapt the layerwise probing methodology often employed for linguistic pattern detection by 
~\citet{vanaken_2019,tenney2019bert,liu2019linguistic} for the probe tasks suggested in~\citet{petroni2019language}.

%% file: setup.tex
\section{Experimental Setup}
\label{sec:setup}

\subsection{Models}
\label{sec:models}

\bert{} is a bidirectional text encoder built by stacking several transformer layers. 
\bert{} is often pre-trained with two tasks: next sentence classification and masked language modeling (MLM). 
MLM is cast as a classification task over all tokens in the vocabulary.
It is realized by training a decoder that takes as input the mask token embedding and outputs a probability distribution over vocabulary tokens. In our experiments we used \bert{} base (12 layers) pretrained on the BooksCorpus ~\cite{zhu2015bookCorpus} and English Wikipedia. 
We use this model for fine-tuning to keep comparisons consistent. 
Henceforth, we refer to pre-trained \bert{} as just \bert{}. The following is a list of all fine-tuned models used in our experiments:

\begin{enumerate}
    \item \ner: (cased) named entity recognition model tuned on Conll-2003~\cite{tjongkimsang2003conll}. 
    \item \qasquad: A question answering model (span prediction) trained on SQuAD 1~\cite{rajpurkar2016squad}. The trained model achieved an F1 score of 88.5 on the test set. 
    
    \item \qasquadbig: QA span prediction trained Squad 2~\cite{DBLP:journals/corr/abs-1806-03822}. 
    The F1 score was 67 (note: SQUAD 2 is a more challenging version of SQUAD 1).
    
    \item \rankmarco: Ranking model trained on the MSMarco passage reranking task~\cite{bajaj2016ms}. We used the fine-tuning procedure described in \cite{nogueira2019passage} to obtain a regression model that predicts a relevance score given query and passage. 
    
    \item \mlmmarco: \bert{} fine-tuned on the passages from the MSMarco dataset using the masked language modeling objective as per~\cite{devlin2018bert}. 15\% of the tokens masked at random.  
    
    \item \mlmsquad: \bert{} fine-tuned on text from SQUAD using the masked language modeling objective as per~\citet{devlin2018bert}. 15\% of the tokens masked at random.
    
\end{enumerate}

\begin{table*}[]
\centering
\small
\begin{tabular}{lccll}
\hline
Name & \#Rels & \#Instances & Example & Answer \\
\hline
\concpetprobe{} & - & 12514 & Rocks are [MASK]. & {solid} \\
\trexprobe{} & 41 & \ 34017 & The capital of Germany is [MASK]. & {Berlin} \\
\greprobe{} & 3 & 5528 & Eyolf Kleven was born in [MASK]. & {Copenhagen} \\
\squadprobe{} & - & 305 & Nathan Alterman was a [MASK]. & {Poet} \\
\hline
\end{tabular}
\caption{Knowledge probes used in the experiments. \citet{petroni2019language} subsampled \concpetprobe{} \cite{speer2012representing}, \trexprobe{} \cite{elsahar2019t}, \greprobe{} \cite{orr201350} and \squadprobe{} \cite{rajpurkar2016squad}.}
\label{tab:kp_details}
\end{table*}

When fine-tuning, our goal was to not only achieve good performance but also to minimize the number of extra parameters added. More parameters outside \bert{} may increase the chance of knowledge being stored elsewhere leading to unreliable measurement. We used the Huggingface transformers library \cite{Wolf2019HuggingFacesTS} for implementing all models in our experiments. More details on hyperparameters and training can be found in the Appendix.

\subsection{Knowledge probes}
\label{sec:probes}
We utilized the existing suite of LAMA knowledge probes suggested in~\cite{petroni2019language}\footnote{ https://github.com/facebookresearch/LAMA} for our experiments. Table~\ref{tab:kp_details} briefly summarizes the key details. The probes are designed as cloze statements and limited to single token factual knowledge, i.e., multi-word entities and relations are not included.

Each probe in LAMA is constructed to test a specific relation or type of relational knowledge.  \concpetprobe{} is designed to test for general conceptual knowledge since it masks single token objects from randomly sampled sentences whereas \trexprobe{} consists of hundreds of sentences for 41 specific relationship types like \textit{member of} and  \textit{language spoken}. 
\greprobe{} tests for 3 specific types of factual knowledge related to people: place-of-birth (2937), date-of-birth (1825), and place-of-death (766 instances). The date-of-birth is a strict numeric prediction that is not covered by \trexprobe{}. 
Finally, \squadprobe{} uses context insensitive questions from SQuAD that have been manually rewritten to cloze-style statements. 
Note that this is the same dataset used to train \qasquad{} and \qasquadbig{}. 


\subsection{Probing Procedure}
\label{sec:procedure}
Our goal is to measure the knowledge stored in \bert{} via knowledge probes. 
LAMA probes rely on the MLM decoding head to complete cloze statement tasks. Note that this decoder is only trained for the mask token embedding of the final layer and is unsuitable if we want to probe all layers of \bert{}. 
To overcome this we train a new decoding head for each layer of a \bert{} model under investigation.

\mpara{Training}: We train a new decoding head for each layer the same way as standard pre-training using MLM. 
We also used Wikipedia (WikiText-2 data) -- sampling passages at random and then randomly masking 15\% of the tokens in each.
Our decoding head uses the same architecture as proposed by~\citet{devlin2018bert} -- a fully connected layer with GELU activation and layer norm (epsilon of 1e-12) resulting in a new 768 dimensional embedding. This embedding is then fed to a linear layer with softmax activation to output a probability distribution over the 30K vocabulary terms. In total, the decoding head possesses $\sim$24M parameters. We froze \bert{}'s parameters and trained the the decoding head only for every layer using the same training data. We initialized the new decoding heads with the parameters of the pre-trained decoding and then fine-tuned it. Our experiments with random initialization yielded no significant difference in the early and middle layers and worse performance on the last few layers. We used a batch size of 8 and trained until validation loss was minimized using the Adam optimizer~\cite{kingma2014adam}. With the new decoding heads, the LAMA probes can be applied to every layer. 



\mpara{Measuring Knowledge}  We convert the probability distribution output of the decoding head to a ranking with the most probable token at rank 1. The amount of knowledge stored at each layer is measured by precision at rank 1 (P@1 for short). We use P@1 as the main metric in all our experiments. Since rank depth of 1 is a strict metric, we also measured P@10 and P@100. We found the trends to be similar across varying rank depths. For completeness, results for P@10 and P@100 can be found in the appendix. Additionally, we measure the total amount of knowledge contained in \bert{} by 
$$\mathcal{P}@1 = max( \{ P^{l}@1  | \,\, \forall l \in L  \} )$$

where $L$ is the set of all layers and $P^{l}@1$ is the P@1 for a given layer $l$. In our experiments $|L|=12$. This metric allows us to consider knowledge captured at all layers of \bert{}, not just a specific layer. If knowledge is always best captured at one specific layer $l$ then $\mathcal{P}$@1 = $P^{l}@1$. If the last layer always contains the most information then total knowledge is equal to the knowledge stored in the last layer.

\mpara{Caveats of probing with cloze statements}: Note that \bert{}, \mlmmarco{}, and \mlmsquad{} are trained for the task of masked word prediction which is exactly the same task as our probes. The last layers of \bert{} have shown to contain mostly task-specific knowledge -- how to predict the masked word in this case~\cite{kovaleva-etal-2019-revealing}. Hence, good performance in our probes at the last layers for MLM models can be partially attributed to task-based knowledge.
\todo{Mention what classifier you used specifically. JS: i dont understand. which classifier?}

%% file: evolution-new.tex
\section{Results}
\label{sec:results}

In contrast to existing work, we want to analyze relation knowledge across layers to measure the total knowledge contained in \bert{} and observe the evolution of relational knowledge through the layers. 

\begin{figure*}[h!]
    \centering
   \includegraphics[width=0.90\textwidth]{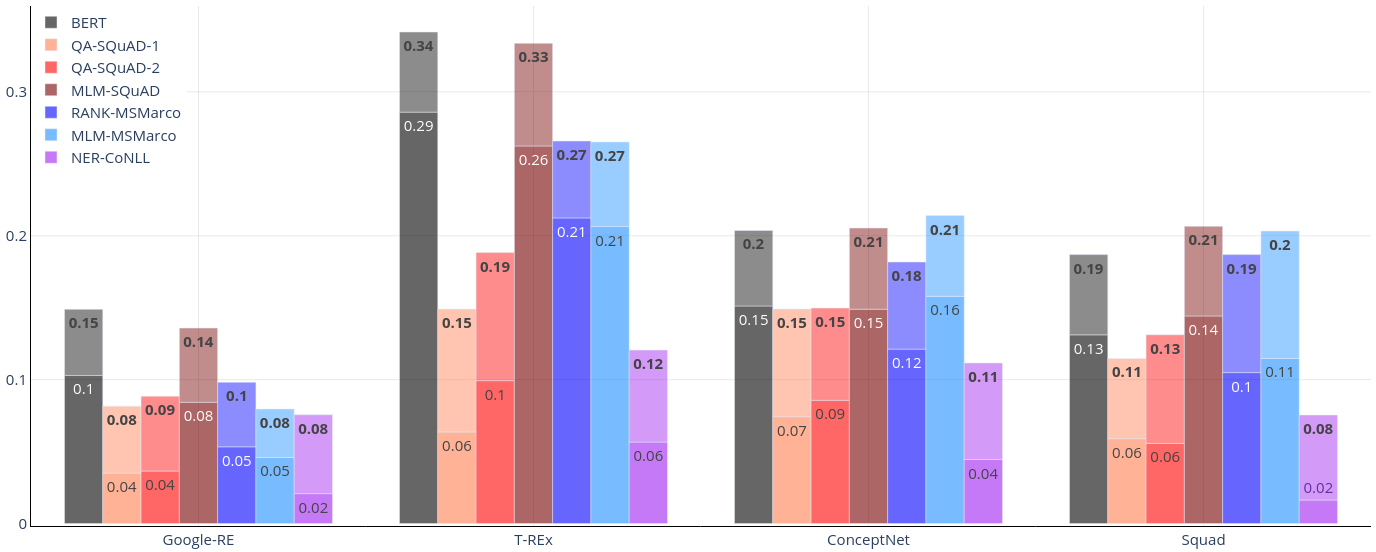}
    \caption{$\mathcal{P}$@k (upper value) vs last layer P@k (lower value) for all models for each LAMA probe.}
    \label{fig:last_layer_vs_union}
\end{figure*}

\todo{update places where you talk about total knowledge but the bar isn't in the graph -- qa squad 1}

\subsection{Intermediate Layers Matter}
\label{sec:intermediate}

The first question we tackle is -- Does knowledge reside strictly in the last layer of \bert{}? 

Figure~\ref{fig:last_layer_vs_union} compares the fraction of correct predictions in the last layer as against all the correct predictions computed at any intermediate layer in terms of $\mathcal{P}@1$.
It is immediately evident that a significant amount of knowledge is stored in the intermediate layers.
While the last layer does contain a reasonable amount of knowledge, a considerable proportion of relations seem to be forgotten and the \textbf{intermediate layers contain relational knowledge that is absent in the final layer.}
Specifically, 18\% for \trexprobe{} and 33\% approximately for the others are forgotten by \bert{}s last layer. 
For instance, the answer to \texttt{Rocky Balboa was born in [MASK]} is correctly predicted as \texttt{Philadelphia} by Layer 10 whereas the rank of \texttt{Philadelphia} in the last layer drops to $26$ for \bert{}.

The intermediary layers also matter for fine-tuned models. Models with high $\mathcal{P}@1$ tend to have a smaller fraction of knowledge of stored in the intermediate layers -- 20\% for \rankmarco{} on \trexprobe{}. In other cases, the amount of knowledge lost in the final layer is more drastic -- $3\times$ for \qasquadbig{} on \greprobe{}.

We also measured the fraction of relationship types in \trexprobe{} that are better captured in the intermediary layers (Table~\ref{tab:win_loss}). On average, 7\% of all relation types in \trexprobe{} are forgotten in the last layer for \bert{}. \rankmarco{} forgets the least amount of relation types (2\%) whereas \qasquad{} forgets the most (43\%) in \trexprobe{}, while also being the least knowledgeable (lowest or second-lowest $\mathcal{P}@1$ in all probes). This is further proof of our claim that \bert{}'s overall capacity can be better estimated by probing all layers. Surprisingly, \rankmarco{} is able to consistently store nearly all of its knowledge in the last layer. We postulate that for ranking in particular, relational knowledge is a key aspect of the task specific knowledge commonly found in the last layers. 

\subsection{Relational Knowledge Evolution}
\label{sec:evolution}

\begin{figure}[h!]
    \centering
    \includegraphics[width=0.34\textwidth]{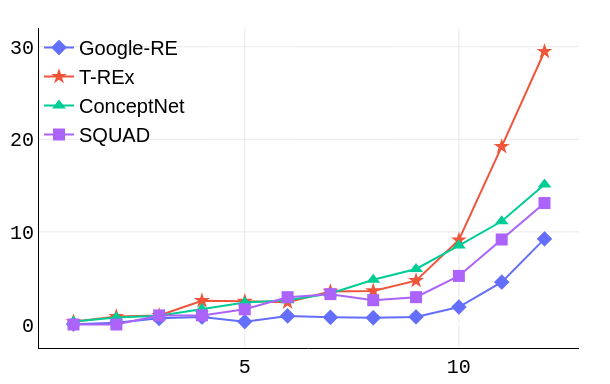}
    \caption{Mean P@1 of \bert{} across all layers.}
    \label{fig:bert_layers}
\end{figure}

Next, we study the evolution of relational knowledge through the \bert{} layers presented in Figure~\ref{fig:bert_layers} that reports P@1 at different layers.

\begin{figure}[h]
    \centering
    \includegraphics[width=0.39\textwidth]{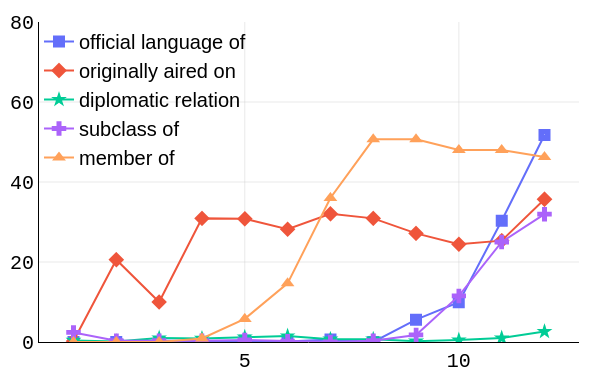}
    \caption{P@1 across all layers for \bert{} for select relationship types from \trexprobe{}.}
    \label{fig:trex_all_default}
\end{figure}

We observe that the amount of \textbf{relational knowledge captured increases steadily with each additional layer}.
While some relations are easier to capture early on, we see an almost-exponential growth of relational knowledge after Layer 8.
This indicates that relational knowledge is predominantly stored in the last few layers as against low-level linguistic patterns are learned at the lower layers (similar to ~\citet{vanaken_2019}).
In Figure~\ref{fig:trex_all_default} we inspect relationship types that show uncharacteristic growth or loss in \trexprobe{}. 

While \texttt{member of} is forgotten in the last layers, the relation \texttt{diplomatic relation} is never learned at all, and  \texttt{official language of} is only identifiable in the last two layers. Note that the majority of relations follow the nearly exponential growth curve of the mean performance in Figure~\ref{fig:bert_layers} (see line \trexprobe{}). From our calculations, nearly 15\% of relationship types double in mean P@1 at the last layer.

We now analyze evolution in fine-tuned models to understand the impact of fine-tuning on the knowledge contained through the layers.
There are two effects at play once \bert{} is fine-tuned. 
First, during fine-tuning \bert{} observes additional task-specific data and hence has either opportunity to monotonically increase its relational knowledge or replace relational knowledge with more task-specific information. Second, the task-specific loss function might be misaligned with the MLM probing task. 
This means that fine-tuning might result in difficulties in retrieving the actual knowledge using the MLM head.
In the following, we first look at the overall results and then focus on specific effects thereafter.

\begin{figure}[h!]
    \centering
    \includegraphics[width=0.43\textwidth]{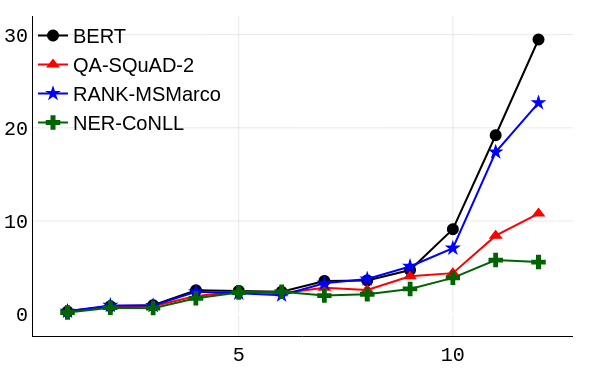}
    \caption{Knowledge contained per layer measured in terms of P@1 on \trexprobe{}.}
    \label{fig:layerwise_mixed}
\end{figure}

Figure~\ref{fig:layerwise_mixed} shows the evolution of knowledge in 3 different models when compared to \bert{}. 

All models possess nearly the same amount of knowledge until layer 6 but then start to grow at different rates. Most surprisingly, \rankmarco{}'s evolution is closest to \bert{} whereas the other models forget information rapidly. With previous studies indicating that the last layers make way for task-specific knowledge
~\cite{kovaleva-etal-2019-revealing}, the ranking model can retain a larger amount of knowledge when compared to other fine-tuning tasks in our experiments.
\todo{Empty cite above}

These results raise the question: Is \rankmarco{} able to retain more knowledge because MSMarco is a bigger dataset or is it because the ranking objective is better suited to knowledge retention as compared to QA, MLM or NER?

\todo{is the previous question answered later. If not we say we leave this for future work.}

%% file: fine-tuned.tex
\subsection{Effect of fine-tuning data}
\label{sec:data_size}


To isolate the effect of the fine-tuning dataset, we first fix the fine-tuning objective. 
We experimented with an MLM and a QA span prediction objective. 
For MLM, we used models trained on fine-tuning task data of varying size -- \bert{}, \mlmmarco{} ($\sim$ 8.8 million unique passages) and \mlmsquad{} ($\sim$ 500+ unique articles). For the QA objective, we experimented with \qasquad{} and \qasquadbig{} which utilize the same dataset of passages but \qasquadbig{} is trained on 50K extra unanswerable questions.

\begin{figure*}
\begin{subfigure}{.22\textwidth}
  \centering
  \includegraphics[width=.99\linewidth]{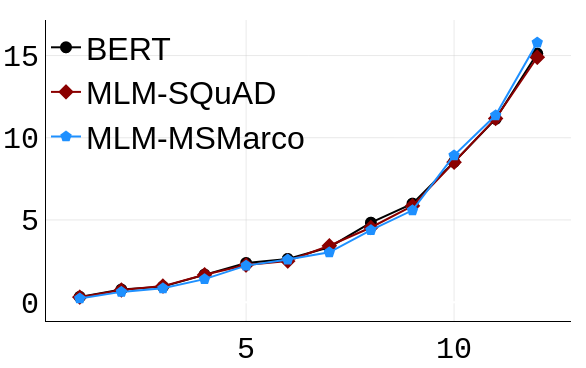}
  \caption{ConceptNet}
  \label{fig:sfig1}
\end{subfigure}%
\begin{subfigure}{.24\textwidth}
  \centering
  \includegraphics[width=.99\linewidth]{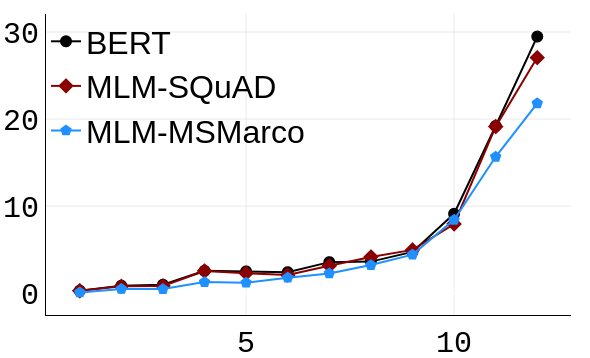}
  \caption{T-REx}
  \label{fig:sfig2}
\end{subfigure}%
\begin{subfigure}{.24\textwidth}
  \centering
  \includegraphics[width=.99\linewidth]{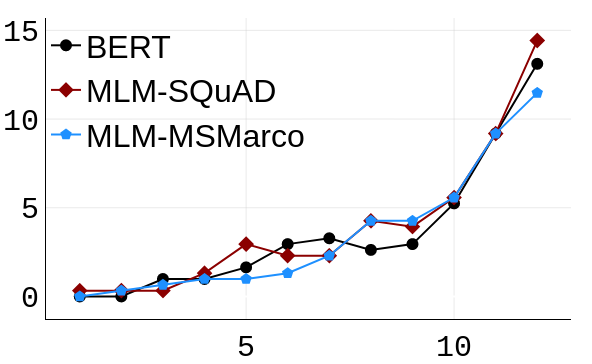}
  \caption{Squad}
  \label{fig:sfig3}
\end{subfigure}
\begin{subfigure}{.24\textwidth}
  \centering
  \includegraphics[width=.99\linewidth]{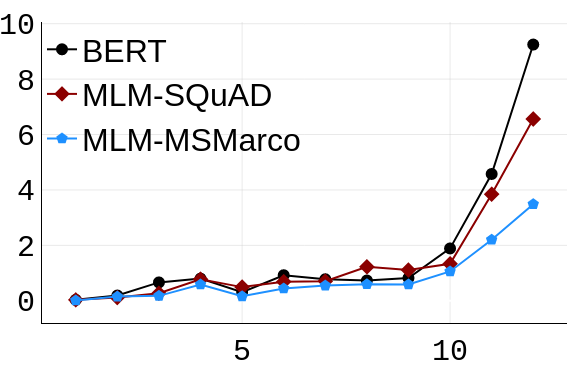}
  \caption{Google-RE}
  \label{fig:sfig3}
\end{subfigure}
\caption{Effect of dataset size. Mean P@1 across layers for \bert{}, \mlmmarco{} and \mlmsquad{}.}
\label{fig:dataset_mlm}
\end{figure*}

\begin{figure*}
\begin{subfigure}{.22\textwidth}
  \centering
  \includegraphics[width=.99\linewidth]{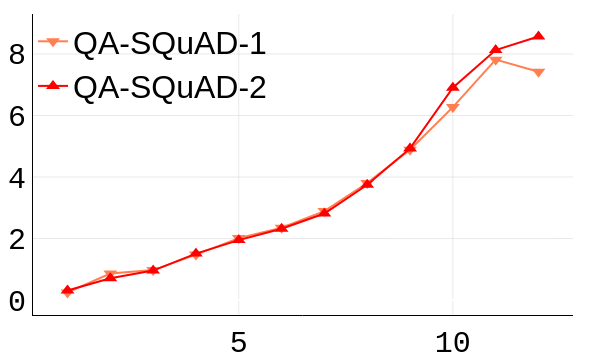}
  \caption{ConceptNet}
  \label{fig:sfig1}
\end{subfigure}
\begin{subfigure}{.24\textwidth}
  \centering
  \includegraphics[width=.99\linewidth]{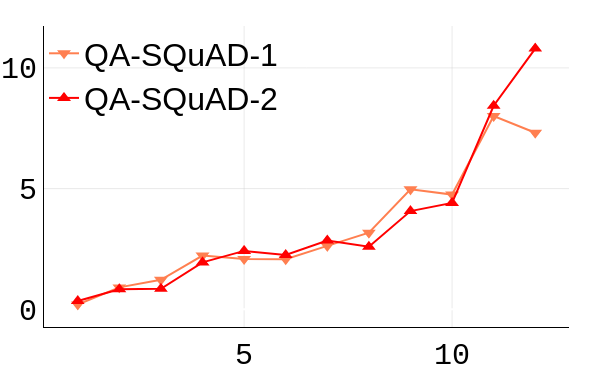}
  \caption{T-REx}
  \label{fig:sfig2}
\end{subfigure}%
\begin{subfigure}{.24\textwidth}
  \centering
  \includegraphics[width=.99\linewidth]{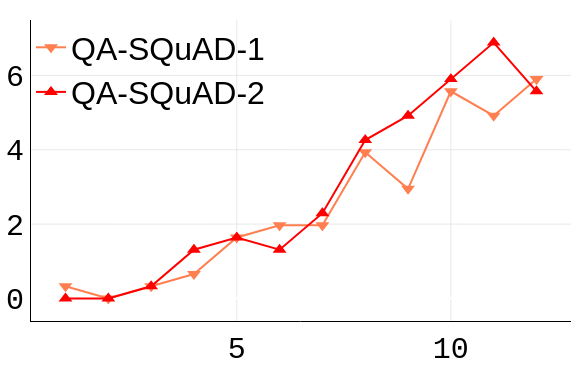}
  \caption{Squad}
  \label{fig:sfig3}
\end{subfigure}
\begin{subfigure}{.24\textwidth}
  \centering
  \includegraphics[width=.99\linewidth]{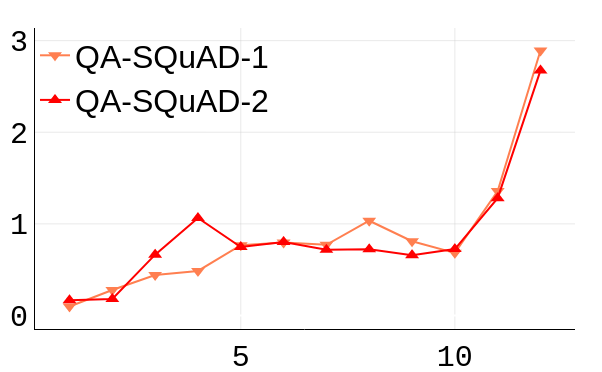}
  \caption{Google-RE}
  \label{fig:sfig3}
\end{subfigure}
    \caption{Effect of dataset size. Mean P@1 across layers for \qasquad{} and \qasquadbig{}. }
    \label{fig:dataset_qa_squad}
\end{figure*}

Figure~\ref{fig:last_layer_vs_union} shows the total knowledge and Figure~\ref{fig:dataset_mlm} shows the evolution of knowledge for both MLM models as compared to \bert{}. When fine-tuning, \bert{} seemingly tends to forget some relational knowledge to accommodate for more domain-specific knowledge. We suspect it forgets certain relations (found in the probe) to make way for other knowledge not detectable by our probes. In the case where the probe is aligned with the fine tuning data (\squadprobe{}), \mlmsquad{} learns more about its domain and outperforms \bert{} but only by a small margin ($<5\%$). Even though \mlmmarco{} uses a different dataset it is able to retain a similar level of knowledge in \squadprobe{}. The evolution trends in Figure~\ref{fig:dataset_mlm} further confirm that fine tuning leads to forgetting mostly in the last layers. Since the fine tuning objective and probing tasks are aligned, it is more evident in these experiments that relational knowledge is being forgotten or replaced.

When observing $\mathcal{P}@1$ and $P@1$, according to \trexprobe{} and \greprobe{} in particular, \mlmmarco{} forgets a large amount of knowledge but retains common sense knowledge (\concpetprobe{}). \mlmsquad{} contains substantially more knowledge overall according to 2/4 probes and nearly the same in the others as compared to \mlmmarco{}. Seemingly, the amount of knowledge contained in fine tuned models is not directly correlated with the size of the dataset. There can be several contributing factors to this phenomenon potentially related to the data distribution and alignment of the probes with the fine tuning data. We leave these avenues open to future work.

Considering the QA span prediction objective, we first see that the total amount of knowledge stored ($\mathcal{P}@1$) in \qasquadbig{} is higher for 3/4 knowledge probes (from Figure
~\ref{fig:last_layer_vs_union}). Figure~\ref{fig:dataset_qa_squad} shows the evolution of knowledge captured for \qasquad{} vs \qasquadbig{}. \qasquadbig{} captures more knowledge at the last layer in 3/4 probes with both models showing similar knowledge emergence trends. This result hints to the fact that a more difficult task (SQUAD2) on the same dataset forces BERT to remember more relational knowledge in its final layers as compared to the relatively simpler SQUAD1. This point is further emphasized in Table~\ref{tab:win_loss}. Only 17\% of relation types are better captured in the intermediary layers of \qasquadbig{} as compared to 43\% for \qasquad{}. 

\begin{table}[]
\centering
\small
\begin{tabular}{lccc}
\hline
\textbf{Models}                      & \textbf{P@1} & \textbf{P@10} & \textbf{P@100} \\ \hline
\multicolumn{1}{l|}{\bert{}}         & 0.07        & 0.02          & 0.07          \\
\multicolumn{1}{l|}{\qasquad{}}    & 0.43        & 0.38         & 0.38          \\
\multicolumn{1}{l|}{\qasquadbig{}}    & 0.17        & 0.19         & 0.17          \\
\multicolumn{1}{l|}{\mlmsquad{}}   & 0.12        & 0.07         & 0.07          \\
\multicolumn{1}{l|}{\rankmarco{}} & 0.02         & 0.05          & 0.05           \\
\multicolumn{1}{l|}{\mlmmarco{}}     & 0.10        & 0.10         & 0.14          \\
\multicolumn{1}{l|}{\ner}  & 0.26        & 0.33         & 0.43          \\ \hline
\end{tabular}
\caption{\label{tab:win_loss} Fraction of relationship types (of the 41 \trexprobe{}) that are forgotten in the last layer. If mean $P^{12}@1 <$ mean $P^{l}@1$ for a particular relation type then that relation is considered to be forgotten at the last layer.}
\end{table}

\begin{figure*}
\begin{subfigure}{.24\textwidth}
  \centering
  \includegraphics[width=.99\linewidth]{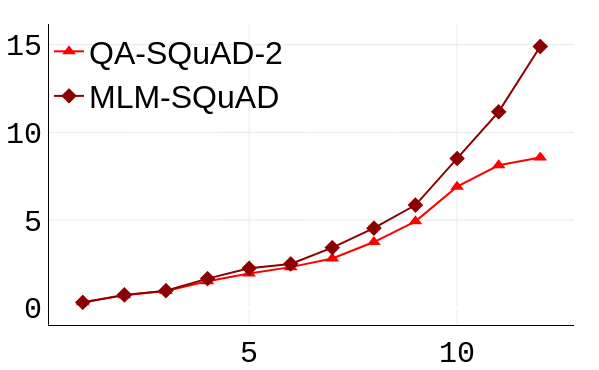}
  \caption{ConceptNet}
  \label{fig:sfig1}
\end{subfigure}%
\begin{subfigure}{.24\textwidth}
  \centering
  \includegraphics[width=.99\linewidth]{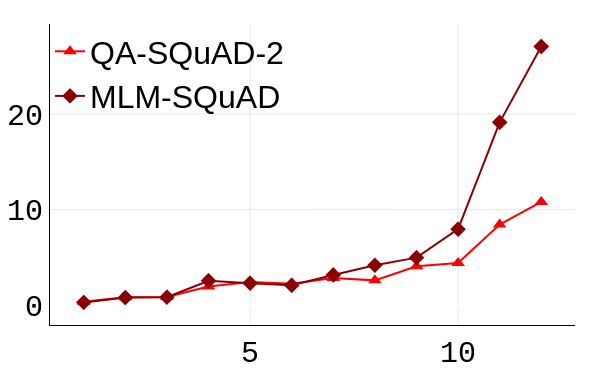}
  \caption{T-REx}
  \label{fig:sfig2}
\end{subfigure}%
\begin{subfigure}{.24\textwidth}
  \centering
  \includegraphics[width=.99\linewidth]{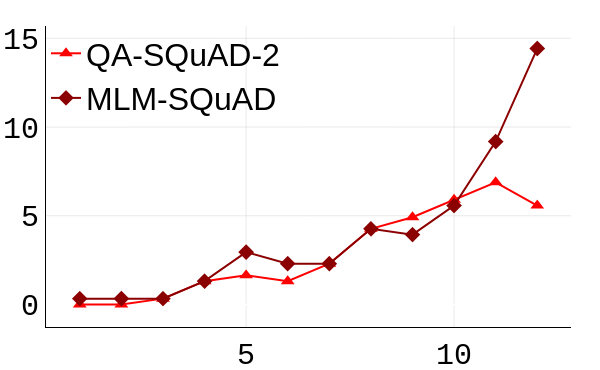}
  \caption{Squad}
  \label{fig:sfig3}
\end{subfigure}
\begin{subfigure}{.24\textwidth}
  \centering
  \includegraphics[width=.99\linewidth]{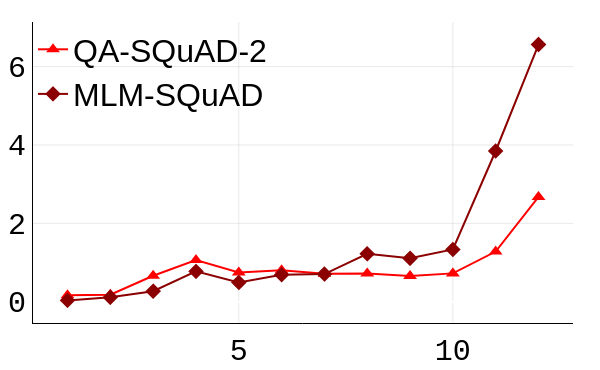}
  \caption{Google-RE}
  \label{fig:sfig3}
\end{subfigure}
    \caption{Effect of Fine-Tuning Objective on fixed size data: SQUAD.}
    \label{fig:finetuning_squad}
\end{figure*}

\begin{figure*}
\begin{subfigure}{.24\textwidth}
  \centering
  \includegraphics[width=.99\linewidth]{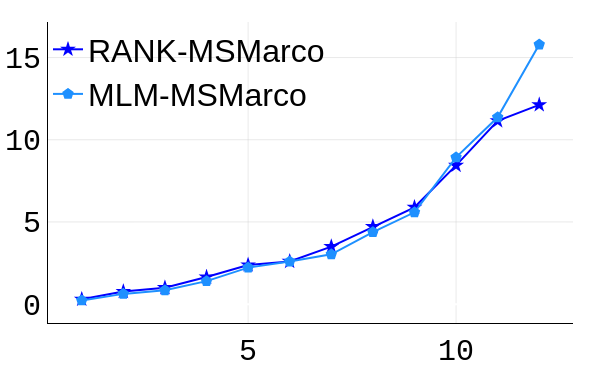}
  \caption{ConceptNet}
  \label{fig:sfig1}
\end{subfigure}%
\begin{subfigure}{.24\textwidth}
  \centering
  \includegraphics[width=.99\linewidth]{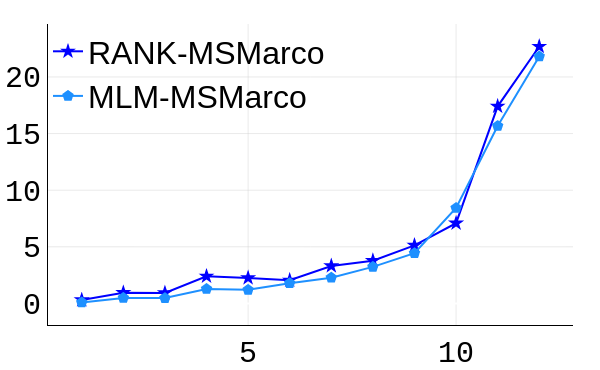}
  \caption{T-REx}
  \label{fig:sfig2}
\end{subfigure}%
\begin{subfigure}{.24\textwidth}
  \centering
  \includegraphics[width=.99\linewidth]{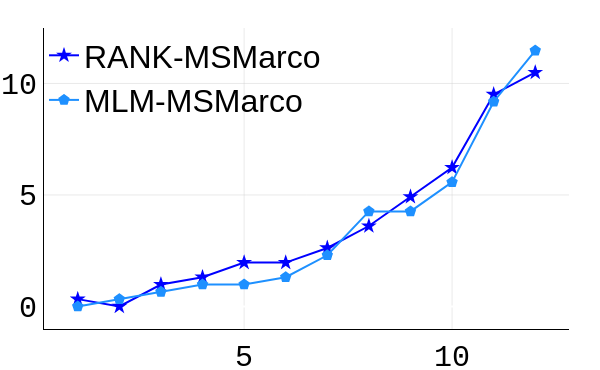}
  \caption{Squad}
  \label{fig:sfig3}
\end{subfigure}
\begin{subfigure}{.24\textwidth}
  \centering
  \includegraphics[width=.99\linewidth]{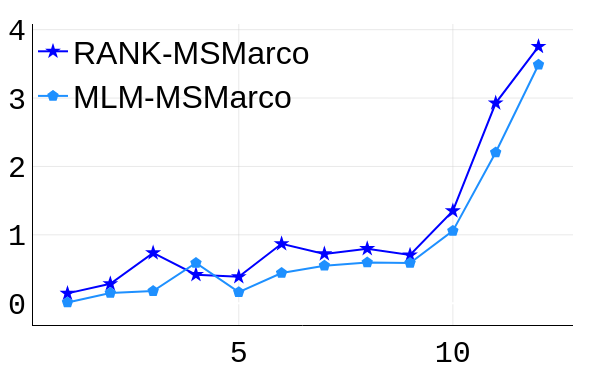}
  \caption{Google-RE}
  \label{fig:sfig3}
\end{subfigure}
    \caption{Effect of Fine-Tuning Objective on fixed size data: MSMarco.}
    \label{fig:finetuning_msm}
\end{figure*}

\subsection{Effect of fine tuning objective}
\label{sec:finetuning}

The second effect that we previously discussed is the effect of the task objective function that might be misaligned with the probing procedure.
To study this effect, we conducted 2 experiments where we fixed the dataset and compared the MLM objective (\mlmmarco{}) vs the ranking objective \rankmarco{} and \mlmsquad{} vs the span prediction objective (\qasquadbig{}). Figure~\ref{fig:finetuning_msm} shows the evolution of knowledge captured for \mlmmarco{} vs \rankmarco{}. 

We observe that \rankmarco{} performs quite similar to \mlmmarco{} across all probes and layers. Although \mlmmarco{} has the same training objective as the probe, the ranking model can retain nearly the same amount of knowledge. We hypothesize that this is because the downstream fine-tuning task is sensitive to relational information.
Specifically, ranking passages for open-domain QA is a task that relies heavily on identifying pieces of knowledge that are strongly related -- For example, given the query: \textit{How do you mow the lawn?}, \rankmarco{} must effectively identify concepts and relations in candidate passages that are related to lawn mowing (like types of grass and lawnmowers) to estimate relevance. 

Reading comprehension /span prediction (QA) however seems to be a less knowledge-intensive task both in terms of total knowledge and at the last layer (Figure~\ref{fig:last_layer_vs_union}). In Figure~\ref{fig:finetuning_squad} we see that the final layers are the most impacted here as well. From Table~\ref{tab:win_loss} we observe that \mlmsquad{} forgets less in its final layer (12\% vs 17\%), with \qasquadbig{} seemingly forgoing relational knowledge for span prediction task knowledge.

%% file: conclusion.tex
\section{Discussion and Conclusion}

In this paper, we introduce a framework to probe all layers of \bert{} for knowledge.
We experimented on a variety of probes and fine-tuning tasks and found that \bert{} contains more knowledge than was reported earlier. Our experiments shed light on the hidden knowledge stored in \bert{} and also some important implications to model building. 
Since intermediate layers contain knowledge that is forgotten by the final layers to make way for task-specific knowledge, our probing procedure can more accurately characterize the knowledge stored.

We show that factual knowledge, like syntactic and semantic patterns, is also forgotten at the last layers due to fine-tuning. However, the last layer can also make way for more domain specific knowledge when the fine tuning objective is the same as the pretraining objective (MLM) as observed in \squadprobe{}. 
Interestingly, forgetting is not mitigated by larger datasets which potentially contain more factual knowledge (\mlmmarco{} $<$ \mlmsquad{} as measured by $\mathcal{P}$@1). 
Instead, we find that knowledge-intensive tasks like ranking do mitigate forgetting compared to span prediction. Although the fine-tuned models always contain less knowledge, with significant (and expected) forgetting in the last layers, \rankmarco{} remembers relatively more relationship types than \bert{} (2\% vs 7\% forgotten) in its last layer (Table
~\ref{tab:win_loss}). This result can partially explain findings in \citet{chang2019pre} where they found that pretraining \bert{} with inverse cloze tasks aids it's transferability to a retrieval and ranking setting. Essentially, ranking tasks encourage the retention of factual knowledge (as measured by cloze tasks) since they are seemingly required for reasoning between the relative relevance of documents to a query.

Our results have direct implications on the use of \bert{} as a knowledge base. By effectively choosing layers to query and adopting early exiting strategies knowldge base completion can be improved. The performance of \rankmarco{} also warrants further investigation into ranking models with different training objectives -- pointwise (regression) vs pairwise vs listwise. More knowledge-intensive QA models like answer generation models may also show a similar trend as ranking tasks but require investigation. We also believe that our framework is well suited to studying variants of BERT architecture and pretraining methods.

%% file: appendix.tex
\clearpage

\section{Appendix}

\subsection{Models}
\begin{itemize}
\item \bert: Off the shelf "bert-base-uncased" from the huggingface transformers library \cite{Wolf2019HuggingFacesTS}
\item \qasquad: Both SQuAD QA models are trained with the huggingface question answering training script \footnote{ https://github.com/huggingface/transformers}. This adds a span prediction head to the default \bert, I.e. a linear layer that computes logits for the span start and span end. So for a given question and a context, it classifies the indices in in which the answer starts and ends. As a loss function it uses crossentropy. The model was trained on a single GPU. We used the huggingface default training script and standard parameters: 2 epochs, learning rate 3e-5, batch size 12.
\item \qasquadbig: Single GPU, also using huggingface training script with standard parameters. Learning rate was 3e-5, batch size 12, best model after 2 epochs.
\item \mlmsquad: Fine tuned on text from SQUAD using the masked language modeling objective as per~\cite{devlin2018bert}. 15\% of the tokens masked at random. Trained for 4 epochs with LR 5e-5. Single GPU.
\item \rankmarco: Trained as described in \cite{nogueira2019passage}. MSMARCO, 100k iterations with batch size 128 (on a TPUv3-8).
\item \mlmmarco: 15\% of the tokens masked at random. 3 epochs, batch size 8, LR 5e-5. Single gpu.
\end{itemize}

\subsection{Experimental results:}
\begin{itemize}
    \item Computing infrastructure used: Everything can be run in Colab notebook with 12gb of RAM and the standard GPU. The experiments, however, have been run on a 
          computing cluster with 6 nodes. Every node had 4 gtx 1080ti and 128gb RAM. Thus being able to parallize the probing of different layers.
    \item Average runtime: Circa 3 hours per layer (that is training the MLM head and probing the LAMA probes) on a single GPU.
    \item Number of parameters: Since we use standard BERT, the base model + MLM head combined have 110,104,890 parameters. The MLM head itself has 24,459,834 parameters.
    \item Validation performance for test results: Since we probed the data, we could not do validation on it.
    \item Explanation of evaluation metrics used with links to code: It is done in knowledge\_probing/probing/metrics.py. But the one that we use are Precision @ k where we just check if the model predicts the correct token at index $<=$ k (P@k)
\end{itemize}

\subsection{Hyperparameter seach:} 
Not applicable.

\subsection{Datasets:}
\begin{itemize}
    \item Wikitext-2: Used for fine-tuning the MLM head. Subset of the Englisch Wikipedia for long term dependency language modeling. 2,088,628 tokens for training, 217.646 for validation, 245.569 for testing. Vocab size: 33,278 out of vocab: 2.6\% of tokens. It can be downloaded from here: https://www.salesforce.com/products/einstein/ai-research/the-wikitext-dependency-language-modeling-dataset/
    \item LAMA probe data: Can be downloaded from their github: https://github.com/facebookresearch/LAMA . Only used for testing. Consists of: Google-RE: 5528 instances over 3 relations. T-REx: 34017 instances over 41 relations. ConceptNet: 12514 instances. This is not grouped into relations. Squad: 305 instances. Context in-sensitive questions rewritten to cloze-statements. No specific relation either.
    \item SQuAD 1.1: Can be downloaded from here: https://rajpurkar.github.io/SQuAD-explorer/ . 100,000+ question answer pairs based on wikipedia articles. Produced by crowdworkers.
    \item SQuAD 2: Can be downloaded from here: https://rajpurkar.github.io/SQuAD-explorer/ . Combines the 100,000+ question answer pairs with 50,000 unanswerable questions. 
    \item MSMARCO: Can be downlaoded from here: https://microsoft.github.io/msmarco/ . For ranking: Dataset for passage reranking was used. Given 1,000 passages, re-rank by relevance. Dataset contains 8,8m passages. For MLM training: Dataset for QA was used. It consists of over 1m queries and the 8,8m passages. Each query has 10 candidate passages. For MLM, we appended the queries with all candidate passages before feeding into BERT.
\end{itemize}

\newpage

\subsection{Knowledge captured in BERT}
\subsubsection{Intermediate Layers Matter}
\begin{figure*}[h!]
\centering
\begin{subfigure}{.24\textwidth}
  \centering
  \includegraphics[width=.99\linewidth]{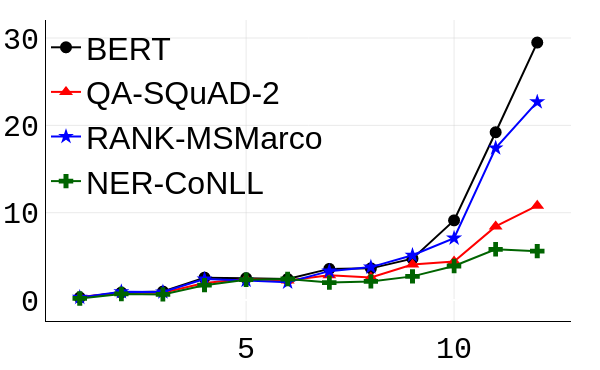}
  \caption{P@1}
  \label{fig:sfig1}
\end{subfigure}%
\begin{subfigure}{.24\textwidth}
  \centering
  \includegraphics[width=.99\linewidth]{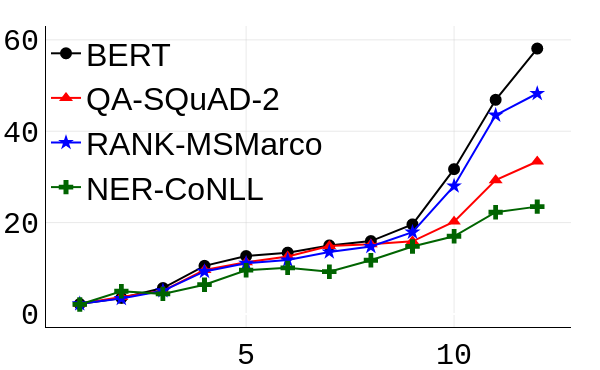}
  \caption{P@10}
  \label{fig:sfig2}
\end{subfigure}%
\begin{subfigure}{.24\textwidth}
  \centering
  \includegraphics[width=.99\linewidth]{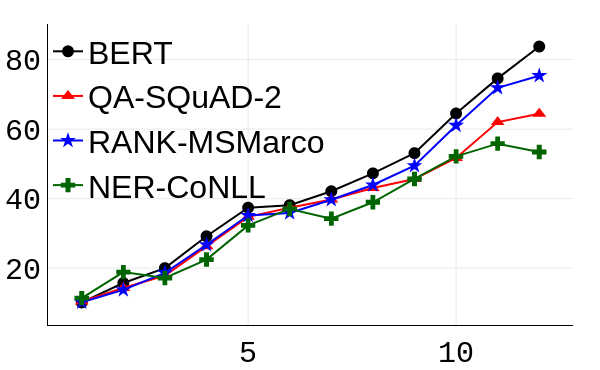}
  \caption{P@100}
  \label{fig:sfig3}
\end{subfigure}
\caption{Mean performance in different precisions on \trexprobe{} sets for \bert, \qasquadbig, \rankmarco, \ner.}
\label{fig:layerwise_mixes_additional_precisions}
\end{figure*}

Additional precisions for Figure \ref{fig:layerwise_mixed} can be found in Figure \ref{fig:layerwise_mixes_additional_precisions}.

\subsubsection{Relational Knowledge Evolution}
\begin{figure*}[h]
\centering
\begin{subfigure}{.24\textwidth}
  \centering
  \includegraphics[width=.99\linewidth]{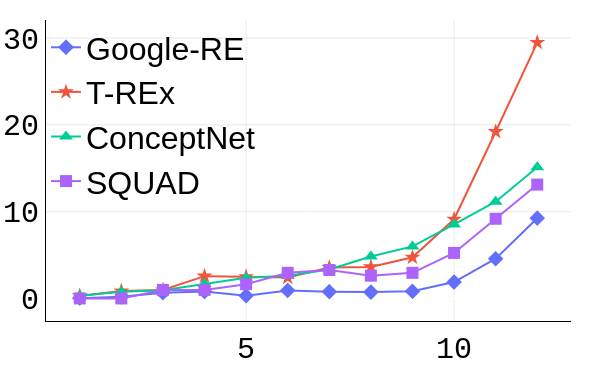}
  \caption{P@1}
  \label{fig:sfig1}
\end{subfigure}%
\begin{subfigure}{.24\textwidth}
  \centering
  \includegraphics[width=.99\linewidth]{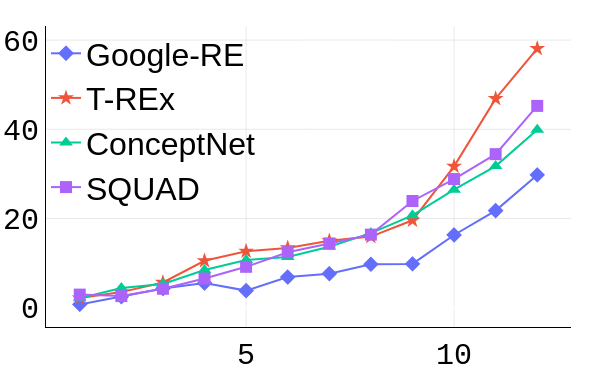}
  \caption{P@10}
  \label{fig:sfig2}
\end{subfigure}%
\begin{subfigure}{.24\textwidth}
  \centering
  \includegraphics[width=.99\linewidth]{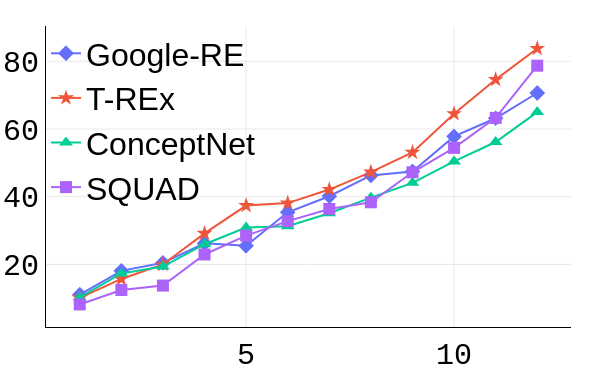}
  \caption{P@100}
  \label{fig:sfig3}
\end{subfigure}
\caption{Mean  performance of \bert{} across all layers and probe sets.}
\label{fig:bert_layers_additional_precisions}
\end{figure*}

Additional precisions for Figure \ref{fig:bert_layers} can be found in Figure \ref{fig:bert_layers_additional_precisions}.

\subsubsection{Effect of dataset size}
\begin{figure*}
\begin{subfigure}{.24\textwidth}
  \centering
  \includegraphics[width=.99\linewidth]{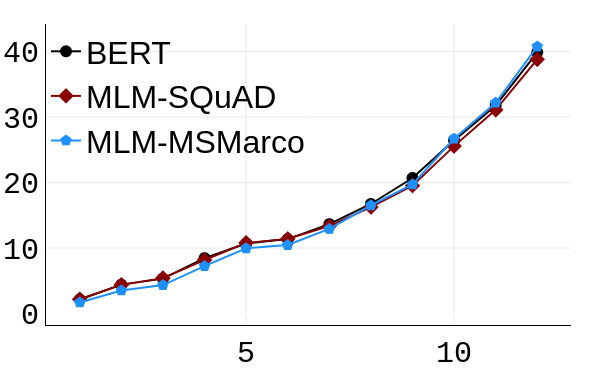}
  \caption{ConceptNet}
  \label{fig:sfig1}
\end{subfigure}%
\begin{subfigure}{.24\textwidth}
  \centering
  \includegraphics[width=.99\linewidth]{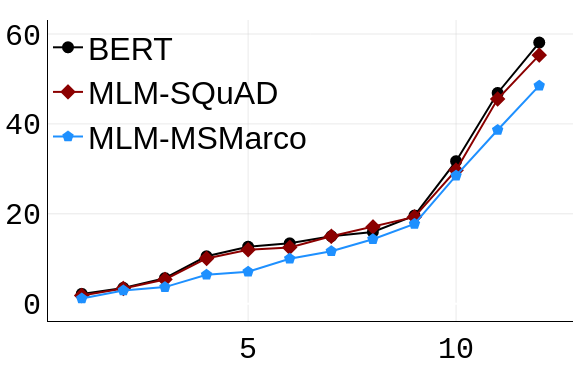}
  \caption{T-REx}
  \label{fig:sfig2}
\end{subfigure}%
\begin{subfigure}{.24\textwidth}
  \centering
  \includegraphics[width=.99\linewidth]{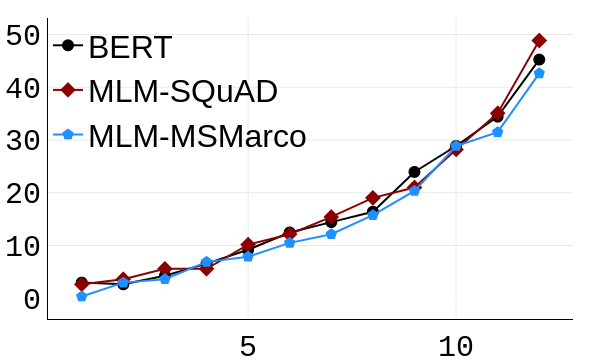}
  \caption{Squad}
  \label{fig:sfig3}
\end{subfigure}
\begin{subfigure}{.24\textwidth}
  \centering
  \includegraphics[width=.99\linewidth]{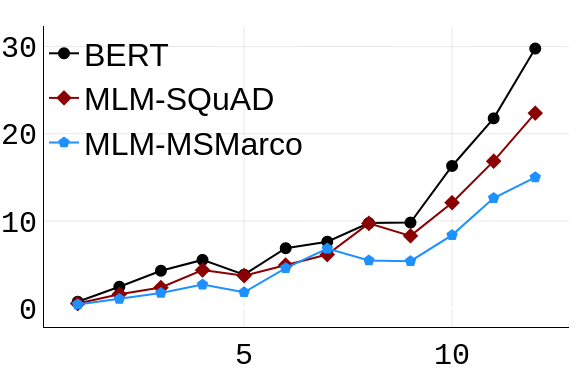}
  \caption{Google-RE}
  \label{fig:sfig3}
\end{subfigure}
\caption{Effect of dataset size. Showing P@10}
\label{fig:dataset_mlm_p_10}
\end{figure*}

\begin{figure*}
\begin{subfigure}{.24\textwidth}
  \centering
  \includegraphics[width=.99\linewidth]{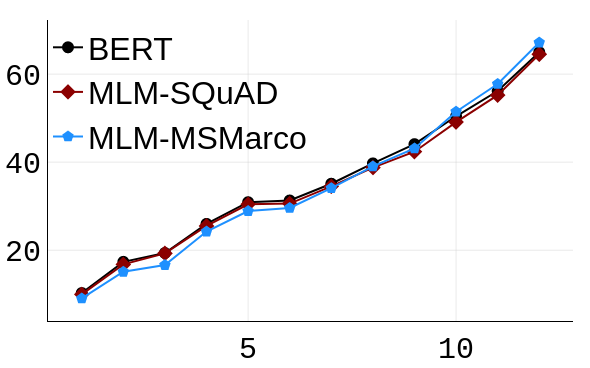}
  \caption{ConceptNet}
  \label{fig:sfig1}
\end{subfigure}%
\begin{subfigure}{.24\textwidth}
  \centering
  \includegraphics[width=.99\linewidth]{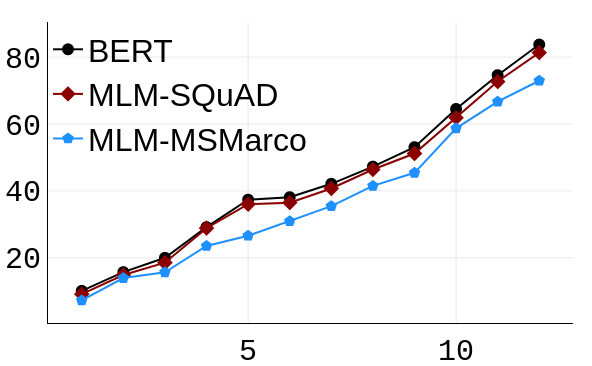}
  \caption{T-REx}
  \label{fig:sfig2}
\end{subfigure}%
\begin{subfigure}{.24\textwidth}
  \centering
  \includegraphics[width=.99\linewidth]{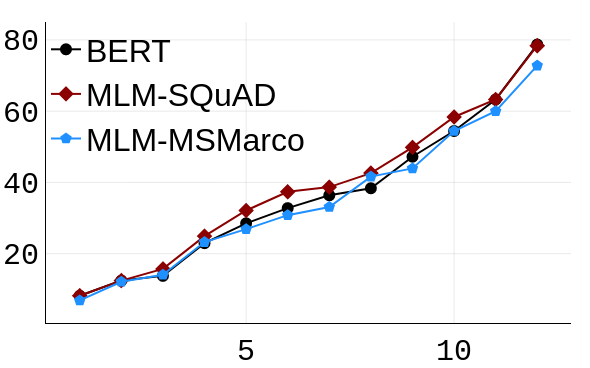}
  \caption{Squad}
  \label{fig:sfig3}
\end{subfigure}
\begin{subfigure}{.24\textwidth}
  \centering
  \includegraphics[width=.99\linewidth]{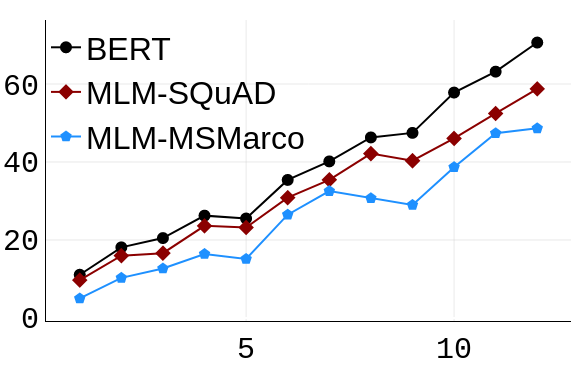}
  \caption{Google-RE}
  \label{fig:sfig3}
\end{subfigure}
\caption{Effect of dataset size. Showing P@100}
\label{fig:dataset_mlm_p_100}
\end{figure*}

\begin{figure*}
\begin{subfigure}{.24\textwidth}
  \centering
  \includegraphics[width=.99\linewidth]{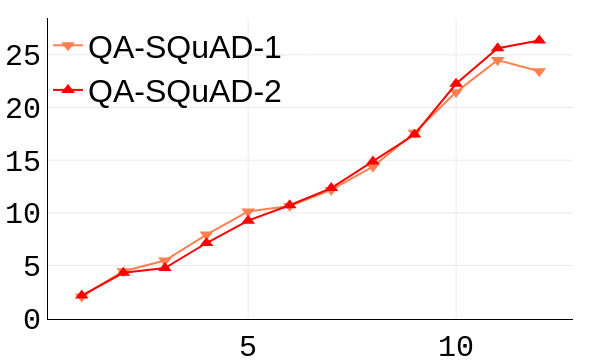}
  \caption{ConceptNet}
  \label{fig:sfig1}
\end{subfigure}%
\begin{subfigure}{.24\textwidth}
  \centering
  \includegraphics[width=.99\linewidth]{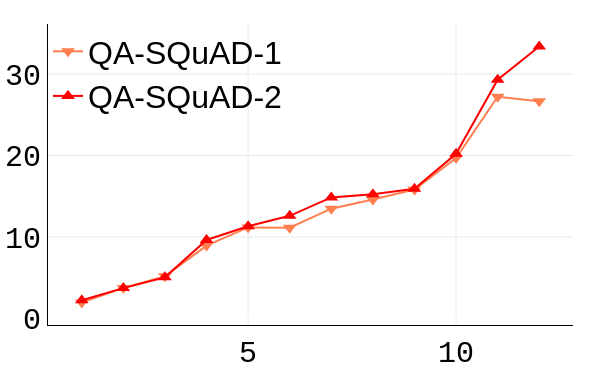}
  \caption{T-REx}
  \label{fig:sfig2}
\end{subfigure}%
\begin{subfigure}{.24\textwidth}
  \centering
  \includegraphics[width=.99\linewidth]{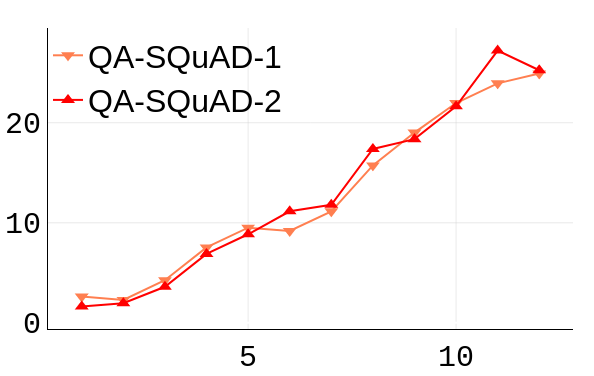}
  \caption{Squad}
  \label{fig:sfig3}
\end{subfigure}
\begin{subfigure}{.24\textwidth}
  \centering
  \includegraphics[width=.99\linewidth]{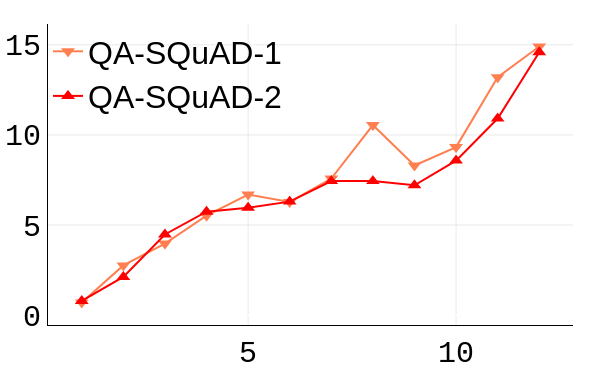}
  \caption{Google-RE}
  \label{fig:sfig3}
\end{subfigure}
    \caption{Effect of dataset size. Showing P@10 for the QA objective.}
    \label{fig:dataset_qa_squad_p_10}
\end{figure*}

\begin{figure*}
\begin{subfigure}{.24\textwidth}
  \centering
  \includegraphics[width=.99\linewidth]{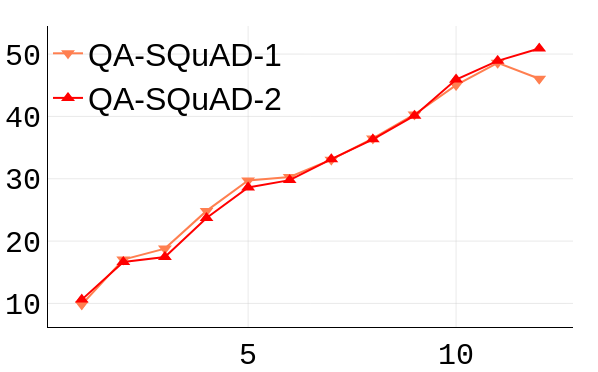}
  \caption{ConceptNet}
  \label{fig:sfig1}
\end{subfigure}%
\begin{subfigure}{.24\textwidth}
  \centering
  \includegraphics[width=.99\linewidth]{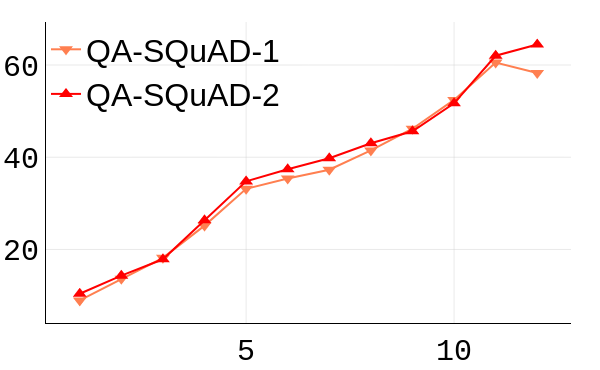}
  \caption{T-REx}
  \label{fig:sfig2}
\end{subfigure}%
\begin{subfigure}{.24\textwidth}
  \centering
  \includegraphics[width=.99\linewidth]{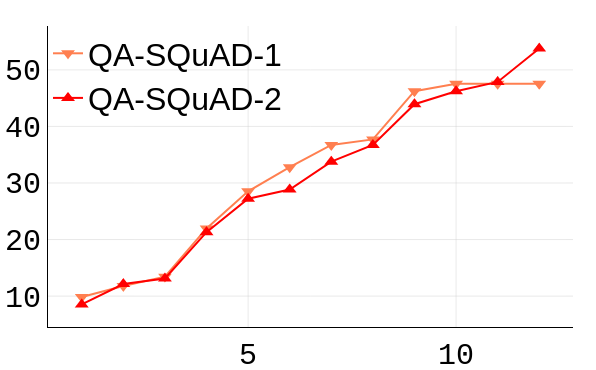}
  \caption{Squad}
  \label{fig:sfig3}
\end{subfigure}
\begin{subfigure}{.24\textwidth}
  \centering
  \includegraphics[width=.99\linewidth]{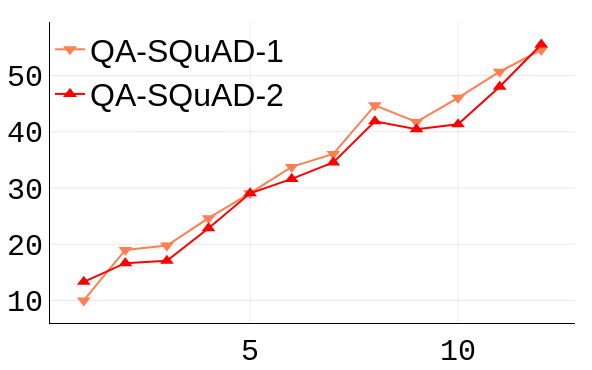}
  \caption{Google-RE}
  \label{fig:sfig3}
\end{subfigure}
    \caption{Effect of dataset size. Showing P@100 for the QA objective.}
    \label{fig:dataset_qa_squad_p_100}
\end{figure*}

Figure \ref{fig:dataset_mlm_p_10} and \ref{fig:dataset_mlm_p_100} show the P@10 and P@100 plots for Figure \ref{fig:dataset_mlm}. Respectively, Figure \ref{fig:dataset_qa_squad_p_10} and \ref{fig:dataset_qa_squad_p_100} show the same for \ref{fig:dataset_qa_squad}.

\subsection{Effect of fine tuning objective}
\begin{figure*}
\begin{subfigure}{.24\textwidth}
  \centering
  \includegraphics[width=.99\linewidth]{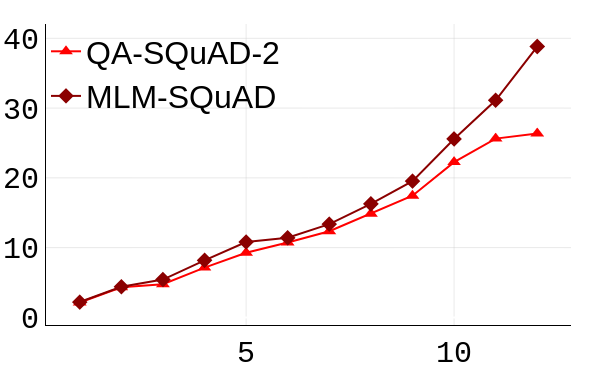}
  \caption{ConceptNet}
  \label{fig:sfig1}
\end{subfigure}%
\begin{subfigure}{.24\textwidth}
  \centering
  \includegraphics[width=.99\linewidth]{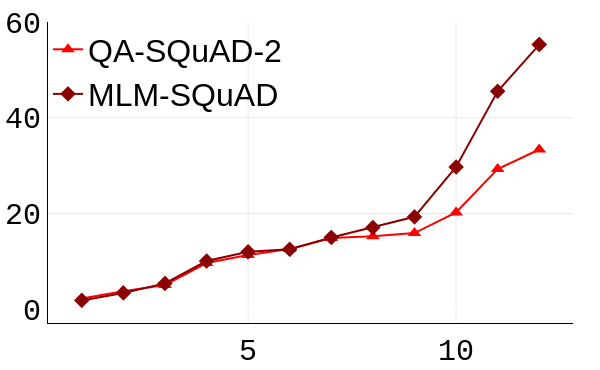}
  \caption{T-REx}
  \label{fig:sfig2}
\end{subfigure}%
\begin{subfigure}{.24\textwidth}
  \centering
  \includegraphics[width=.99\linewidth]{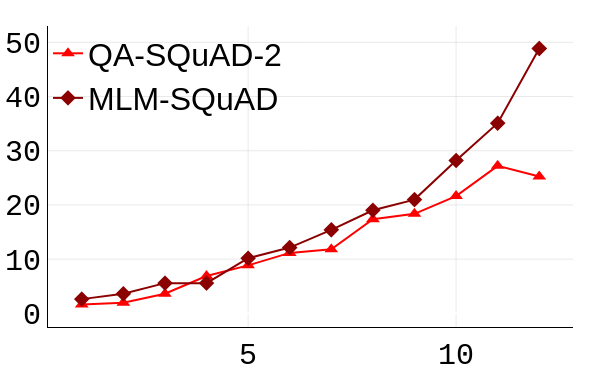}
  \caption{Squad}
  \label{fig:sfig3}
\end{subfigure}
\begin{subfigure}{.24\textwidth}
  \centering
  \includegraphics[width=.99\linewidth]{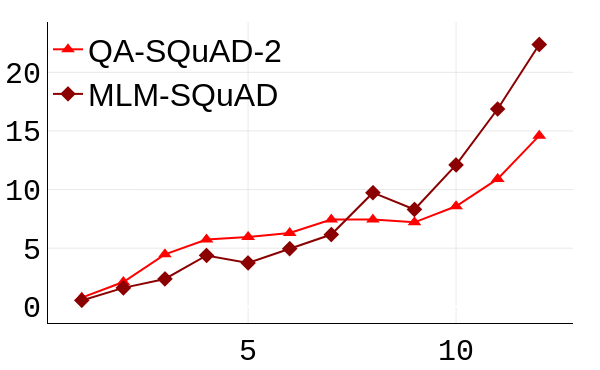}
  \caption{Google-RE}
  \label{fig:sfig3}
\end{subfigure}
    \caption{Effect of Fine-Tuning Objective on fixed size data: SQUAD. Showing P@10.}
    \label{fig:finetuning_squad_p_10}
\end{figure*}

\begin{figure*}
\begin{subfigure}{.24\textwidth}
  \centering
  \includegraphics[width=.99\linewidth]{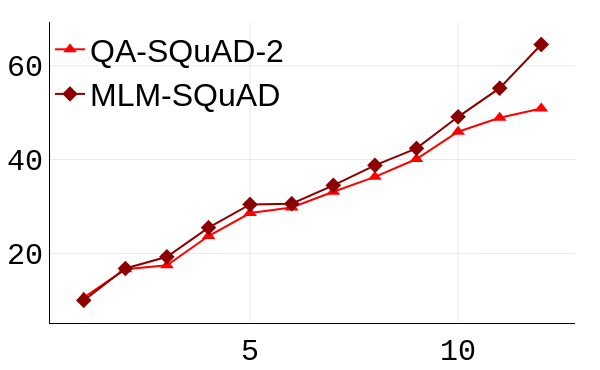}
  \caption{ConceptNet}
  \label{fig:sfig1}
\end{subfigure}%
\begin{subfigure}{.24\textwidth}
  \centering
  \includegraphics[width=.99\linewidth]{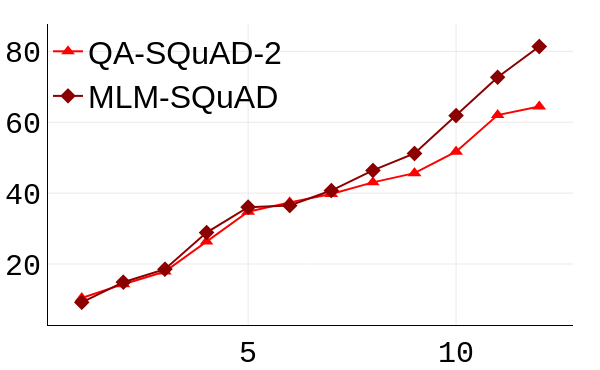}
  \caption{T-REx}
  \label{fig:sfig2}
\end{subfigure}%
\begin{subfigure}{.24\textwidth}
  \centering
  \includegraphics[width=.99\linewidth]{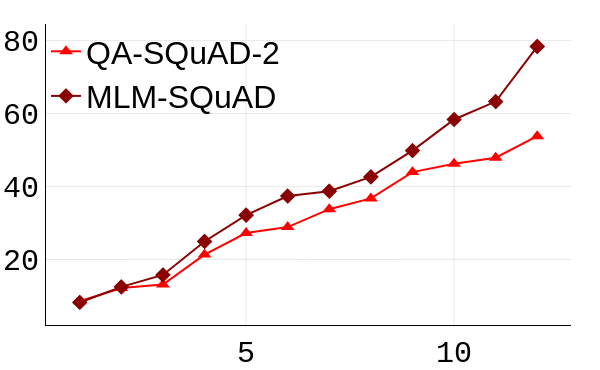}
  \caption{Squad}
  \label{fig:sfig3}
\end{subfigure}
\begin{subfigure}{.24\textwidth}
  \centering
  \includegraphics[width=.99\linewidth]{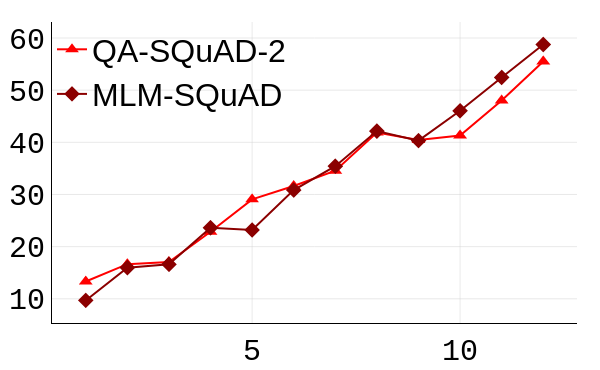}
  \caption{Google-RE}
  \label{fig:sfig3}
\end{subfigure}
    \caption{Effect of Fine-Tuning Objective on fixed size data: SQUAD. Showing P@100.}
    \label{fig:finetuning_squad_p_100}
\end{figure*}

\begin{figure*}
\begin{subfigure}{.24\textwidth}
  \centering
  \includegraphics[width=.99\linewidth]{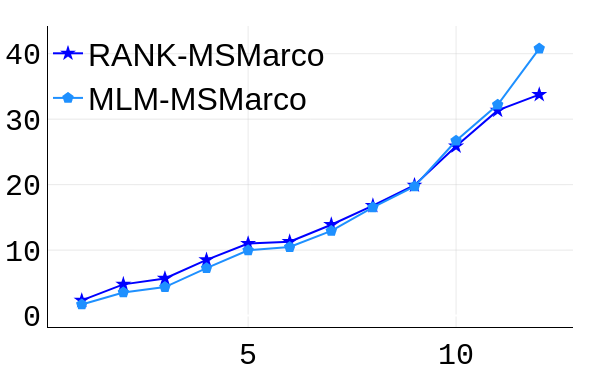}
  \caption{ConceptNet}
  \label{fig:sfig1}
\end{subfigure}%
\begin{subfigure}{.24\textwidth}
  \centering
  \includegraphics[width=.99\linewidth]{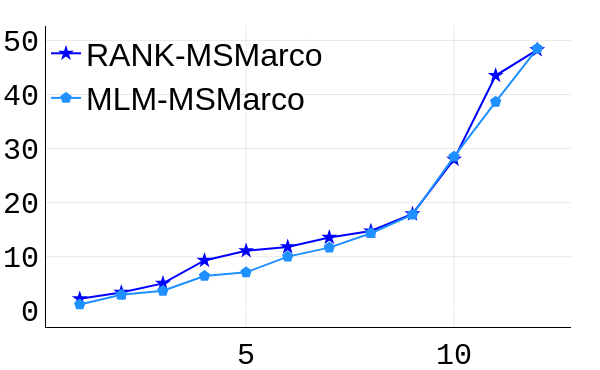}
  \caption{T-REx}
  \label{fig:sfig2}
\end{subfigure}%
\begin{subfigure}{.24\textwidth}
  \centering
  \includegraphics[width=.99\linewidth]{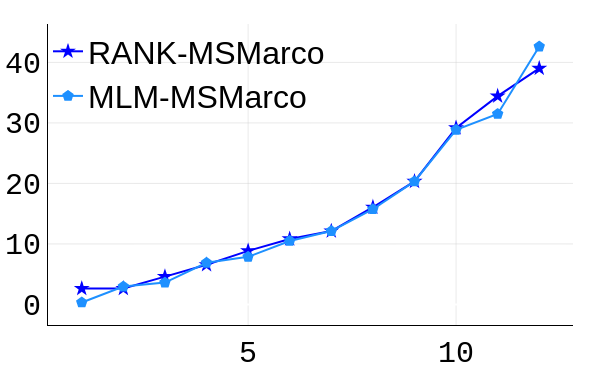}
  \caption{Squad}
  \label{fig:sfig3}
\end{subfigure}
\begin{subfigure}{.24\textwidth}
  \centering
  \includegraphics[width=.99\linewidth]{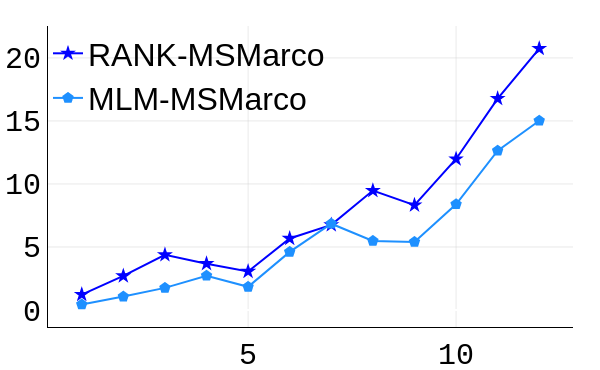}
  \caption{Google-RE}
  \label{fig:sfig3}
\end{subfigure}
    \caption{Effect of Fine-Tuning Objective on fixed size data: MSMarco. Showing P@10.}
    \label{fig:finetuning_msm_p_10}
\end{figure*}

\begin{figure*}
\begin{subfigure}{.24\textwidth}
  \centering
  \includegraphics[width=.99\linewidth]{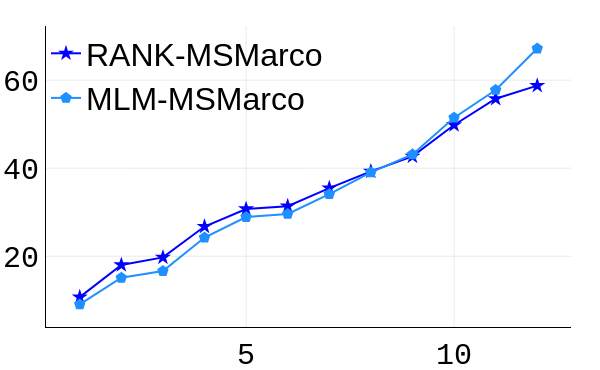}
  \caption{ConceptNet}
  \label{fig:sfig1}
\end{subfigure}%
\begin{subfigure}{.24\textwidth}
  \centering
  \includegraphics[width=.99\linewidth]{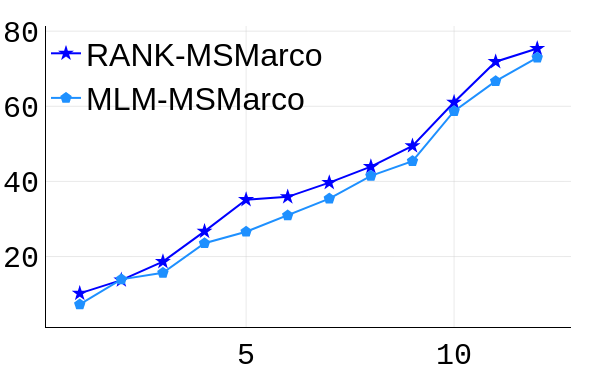}
  \caption{T-REx}
  \label{fig:sfig2}
\end{subfigure}%
\begin{subfigure}{.24\textwidth}
  \centering
  \includegraphics[width=.99\linewidth]{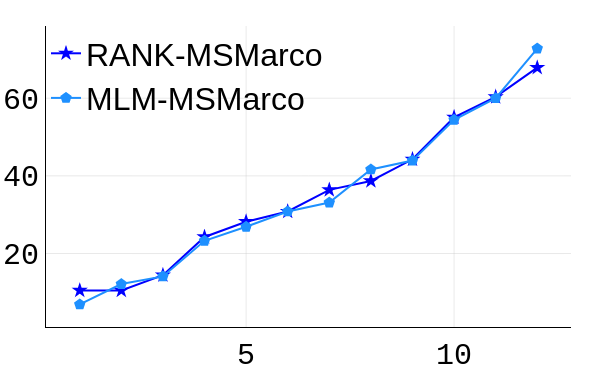}
  \caption{Squad}
  \label{fig:sfig3}
\end{subfigure}
\begin{subfigure}{.24\textwidth}
  \centering
  \includegraphics[width=.99\linewidth]{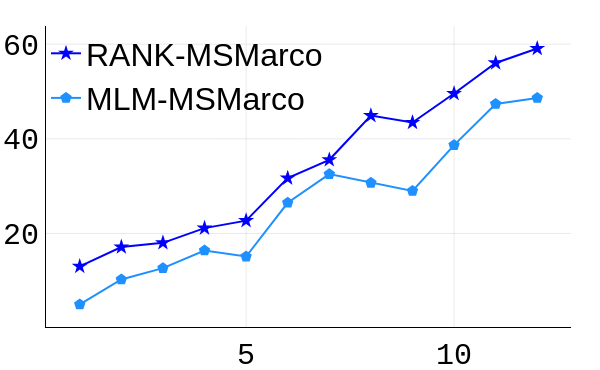}
  \caption{Google-RE}
  \label{fig:sfig3}
\end{subfigure}
    \caption{Effect of Fine-Tuning Objective on fixed size data: MSMarco. Showing P@100.}
    \label{fig:finetuning_msm_p_100}
\end{figure*}

For comparing MLM and QA on SQuAD (\ref{fig:finetuning_squad}), Figure \ref{fig:finetuning_squad_p_10} and \ref{fig:finetuning_squad_p_100} show more precisions. Also, for comparing fine tune objectives on MSMARCO (Figure \ref{fig:finetuning_msm}), Figure \ref{fig:finetuning_msm_p_10} and \ref{fig:finetuning_msm_p_100} show P@10 and P@100. 

\begin{table*}[]
\centering
\begin{tabular}{lcccccccc}
\hline
Model & \multicolumn{2}{c}{\greprobe{}} & \multicolumn{2}{c}{\trexprobe{}} & \multicolumn{2}{c}{\concpetprobe{}} & \multicolumn{2}{c}{\squadprobe{}} \\ \hline
 & P@1 & $\mathcal{P}@1$ & P@1 & $\mathcal{P}@1$ & P@1 & $\mathcal{P}@1$ & P@1 & $\mathcal{P}@1$ \\
\hline
\bert   & 10 & \textbf{15} & 29 & \textbf{34}          & 15 & \textbf{21}               & 13 & \textbf{20}           \\
\qasquad & 3 & \textbf{9}  & 6 & \textbf{15}           & 7 & \textbf{15}   & 5 & \textbf{15}           \\
\qasquadbig     & 3 & \textbf{9} & 10 & \textbf{19} & 8 & \textbf{16} & 6 & \textbf{13}           \\
\mlmsquad      & 4 & \textbf{10} & 15 & \textbf{23}  & 9 & \textbf{16}  & 6 & \textbf{16}           \\
\rankmarco   & 6 & \textbf{11}  & 23 & \textbf{29}          & 12 & \textbf{19} & 10 & \textbf{20}          \\
\mlmmarco    & 3 & \textbf{7} & 14 & \textbf{21}          & 11 & \textbf{17} & 7 & \textbf{12}           \\ \hline
\end{tabular}
\caption{\small{Mean knowledge contained in the last layer (P@1) vs knowledge contained in all layers ($\mathcal{P}@1$) for each probe.} }
\label{tab:last_vs_total}
\end{table*}